\documentclass[runningheads]{llncs}
\usepackage[T1]{fontenc}
\usepackage{graphicx}
\usepackage{xcolor}

\usepackage{hyperref}
\usepackage{amsmath,amssymb,amsfonts}
\usepackage{algorithm}
\usepackage{algorithmic}
\usepackage{makecell}
\usepackage{threeparttable}
\usepackage{graphicx}
\usepackage{textcomp}
\usepackage{xcolor}
\usepackage{multirow}
\usepackage{dsfont}
\usepackage{url}
\usepackage{makecell}
\usepackage{threeparttable}
\usepackage{textcomp}
\usepackage{dsfont}
\usepackage[utf8]{inputenc} 
\usepackage[T1]{fontenc}    
\usepackage{hyperref}       
\usepackage{url}            
\usepackage{booktabs}       
\usepackage{amsfonts}       
\usepackage{nicefrac}       
\usepackage{microtype}      
\usepackage{graphicx}
\usepackage{xcolor}
\usepackage{colortbl,booktabs}
\usepackage{graphicx}
\usepackage{epstopdf}
\usepackage{blindtext, graphicx, mathrsfs, amsmath, amsfonts}
\usepackage{booktabs}
\usepackage{multirow, minipage-marginpar}
\usepackage{caption}
\usepackage{subfigure}
\usepackage{color,soul}
\usepackage[textsize=tiny]{todonotes}
\usepackage{xcolor}
\usepackage{algorithm}
\usepackage{algorithmic}
\usepackage{todonotes}
\usepackage{wrapfig,blindtext}
\begin{document}
\title{Imputing Brain Measurements Across Data Sets via Graph Neural Networks}
%
\titlerunning{Demographic Aware Graph-based Imputation}
%
\author{Yixin Wang\inst{1} \and 
Wei Peng \inst{2}\and 
Susan F. Tapert \inst{3} \and
Qingyu Zhao\inst{2} \and\\
Kilian M. Pohl\inst{2,4}\thanks{Corresponding author: \email{kpohl@stanford.edu} }}
%
\authorrunning{Y. Wang et al.}

\institute{Department of Bioengineering, Stanford University, Stanford, CA, USA
\and  Dept. of Psychiatry \& Behavioral Sciences, Stanford University, Stanford, CA, USA 
\and Department of Psychiatry, University of California, San Diego, CA, USA 
\and Center for Biomedical Sciences, SRI International, Menlo Park, CA, USA 
}
%
%
%
\maketitle              
\begin{abstract} 
Publicly available data sets of structural MRIs might not contain specific measurements of brain Regions of Interests (ROIs) that are important for training machine learning models. For example, the curvature scores computed by Freesurfer are not released by the Adolescent Brain Cognitive Development (ABCD) Study. One can address this issue by simply reapplying Freesurfer to the data set. However, this approach is generally computationally and labor intensive (e.g., requiring quality control). An alternative is to impute the missing measurements via a deep learning approach. However, the state-of-the-art is designed to estimate randomly missing values rather than entire measurements. We therefore propose to re-frame the imputation problem as a prediction task on another (public) data set that contains the missing measurements and shares some ROI measurements with the data sets of interest. A deep learning model is then trained to predict the missing measurements from the shared ones and afterwards is applied to the other data sets. Our proposed algorithm models the dependencies between ROI measurements via a graph neural network (GNN) and accounts for demographic differences in brain measurements (e.g. sex) by feeding the graph encoding into a parallel architecture. The architecture simultaneously optimizes a graph decoder to impute values and a classifier in predicting demographic factors. We test the approach, called \textit{D}emographic \textit{A}ware \textit{G}raph-based \textit{I}mputation (\textit{DAGI}), on imputing those missing Freesurfer measurements of ABCD (N=3760; minimum age 12 years) by training the predictor on those publicly released by the National Consortium on Alcohol and Neurodevelopment in Adolescence (NCANDA, N=540). 5-fold cross-validation on NCANDA reveals that the imputed scores are more accurate than those generated by linear regressors and deep learning models. Adding them also to a classifier trained in identifying sex results in higher accuracy than only using those Freesurfer scores provided by ABCD. 
\keywords{Brain measurements  \and Feature imputation \and Graph representation learning.}
\end{abstract}
%
\section{Introduction}
Neuroscience heavily relies on ROI measurements extracted from structural magnetic resonance imaging (MRI) to encode brain anatomy \cite{symms2004review}. However, public releases of brain measurements might not contain those that are important for a specific task. For example, the Freesurfer scores \cite{fischl2012freesurfer} publicly released by the ABCD study do not contain curvature measurements of cortical regions \cite{casey2018adolescent}, which might be useful for identifying sex differences. While one could theoretically reapply the Freesurfer pipeline to generate those missing measurements, it requires substantial computational resources and manual labor, as, for example, the Freesurfer scores from thousands of MRIs would have to be quality controlled. A more efficient solution is to learn to impute missing brain measurements from the existing ones. 

\begin{figure}[t]
\includegraphics[width=\textwidth]{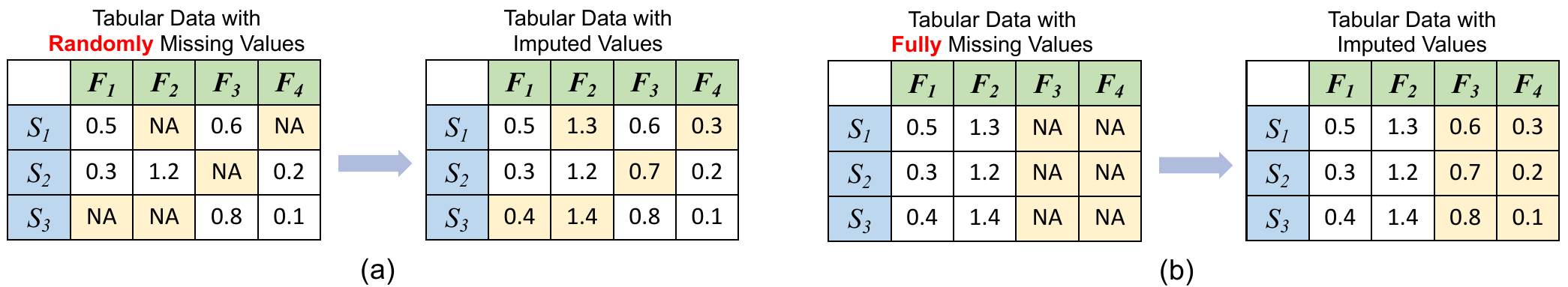}
\caption{Scenarios of missing values : (a) missing values are being randomly distributed across the data set or (b) specific measurements are absent from a data set, which is the problem we aim to solve here.}\label{bg}
\end{figure}

Imputation involves estimating or filling in missing or incomplete data values based on the available data, thereby creating a complete dataset suitable for further analysis or modeling. Examples of popular approaches for imputing measurements are MICE \cite{van2011mice} and k-nearest neighbors \cite{hastie2009elements}. The state-of-the-art in this domain relies on deep learning models, such as using generative autoencoders \cite{talukder2022deep} or graph convolutional networks \cite{vivar2021simultaneous,SPINELLI2020249}. However, such methods assume that missing values are randomly distributed within a matrix capturing all measurements of a data set (refer to Figure \ref{bg} (a)). If each column now represents a measurement, estimating missing values in a column then partly relies on rows (or samples) for which that measurement exists. Here we aim to solve the issue that the entire column does not contain any values (Figure \ref{bg} (b)), i.e., some specific measurements are absent throughout an entire dataset. One could address this issue by combining the data set with the missing values with one that contains them, which then relates to the scenario in Figure \ref{bg} (a). However, the imputation now explicitly depends on the data set with the missing scores so if that data set is updated (e.g., ABCD yearly releases) so do all imputations, which could result in scores conflicting with those imputed based on earlier versions of the data set. We instead address this challenge by re-framing the imputation problem as a prediction task on a single (public) data set, such as NCANDA \cite{ncanda},  that contains the missing measurements and shares some ROI measurements with the data set of interest. A deep learning model can then be trained on NCANDA to predict the curvature scores from the measurements that are shared with ABCD. Afterwards, the trained model is applied to ABCD (or other data sets that share those scores) to predict the missing curvature scores on ABCD. Consequently, our primary objective is to determine the most accurate mapping from the currently available shared measurements to the missing ones. 

Measurements of the same ROI (e.g., cortical thickness and volume) are highly dependent, and measurements of adjacent regions are more likely to be correlated than those from distant regions \cite{mahjoub2018brain,lisowska2019joint}. To explicitly account for such dependencies, our prediction model is based on a graph neural network (GNN) \cite{scarselli2008graph} called Graph Isomorphism Network \cite{xu2018how}. In our graph, each node represents an ROI and adjacent ROIs are connected via edges.  In addition to modeling adjacency of ROIs, our prediction model also accounts for the dependencies between demographic factors and ROI measurements \cite{Llera2019individual,ruigrok2014meta}. For example, women tend to have higher gyrification in frontal and parietal regions than men, which results in the curvature of those ROIs being different between the sexes \cite{luders2006curvature}. We account for this difference by feeding the GNN encodings into a parallel architecture that simultaneously optimizes a graph decoder for imputing values and a classifier for identifying sex. 

We apply our approach, called \textit{D}emographic \textit{A}ware \textit{G}raph-based \textit{I}mputation (\textit{DAGI}), to impute Freesurfer measurements that are available in the NCANDA data set but are missing in ABCD (i.e., ``mean curvature'' and ``Gaussian curvature'')  by explicitly taking advantage of those that are shared among them (i.e., ``average thickness'', ``surface area'' and  ``gray matter volume'' ). Using 5-fold cross-validation, we then show on NCANDA that the accuracy of the imputed scores is significantly higher than those generated by linear regressors and deep learning models. Furthermore, We identify the brain ROIs important in the imputation task by visualizing the learned graph structure via GNNExplainer \cite{ying2019gnnexplainer}. On the ABCD data set, adding the scores to a classifier in identifying sex results in significantly higher accuracy than only using those provided by ABCD or using those imputed by combing the ABCD with the NCANDA data set (Figure \ref{bg} (a)). 

\section{Method}

\begin{figure}[t]
\includegraphics[width=\textwidth]{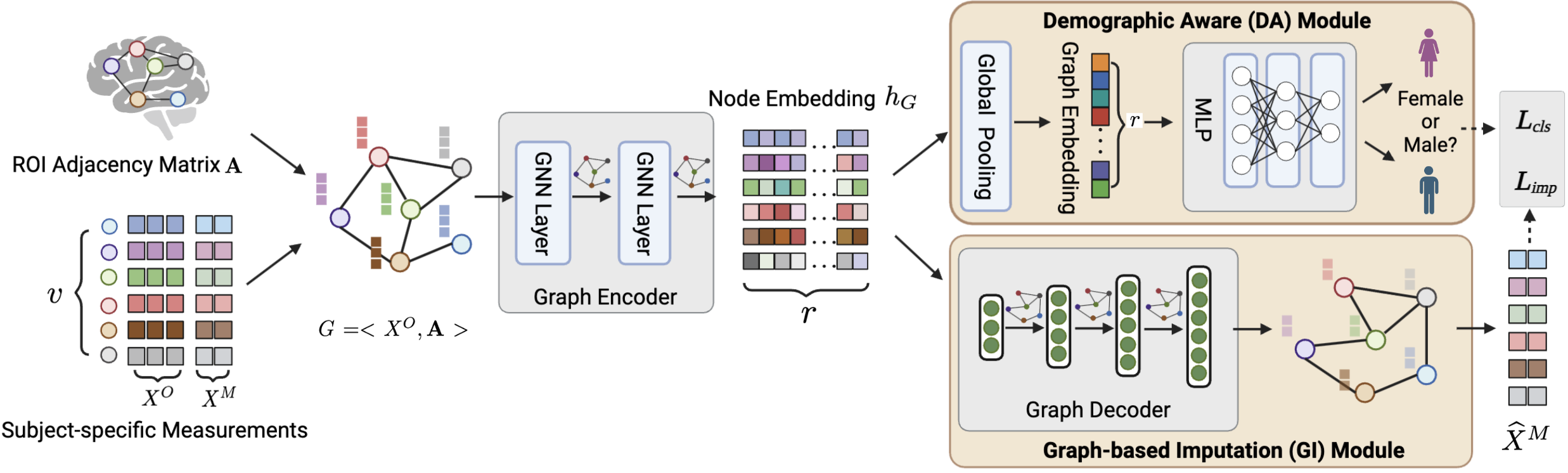}
\caption{Overview of our model: A GNN encodes both adjacency and measurements of brain ROIs into node embeddings, which are utilized by a graph decoder to impute missing values $X^M$. The parallel (upper) branch refines the node representations by differentiating between the sexes.}\label{network}
\end{figure}
Let's assume that the first data set is represented by a matrix $X^1\in \mathbb{R}^{v \times d}$ containing the cortical measurements of $v$ regions, where $d$ cortical measurements $X_i \in \mathbb{R}^{d}$ are extracted from each region $i$. Furthermore, let $X^2\in \mathbb{R}^{v \times p}$ be the data matrix of the second data set, which is based on the same parcellation but contains a different set of measurements for each region, of which $p (<d)$ are those also found in $X^1$. Let $X_i^o \in \mathbb{R}^{1 \times p}$ be the $p$ shared measures across datasets, and $X_i^m  \in \mathbb{R}^{1 \times q}$ be the remaining $q$ measurements only available in $X^1$. Thus, $X^1$ can be divided into $X^O = [X_1^o, ..., X_v^o]^T$ and $X^M = [X_1^m, ..., X_v^m]^T$.
Our goal is to learn an imputation mapping $X^O \rightarrow X^M$ so that we can impute the missing measurements on the second data set. To generate an accurate mapping, we first design a GNN implementation that accounts for dependencies among brain ROI measurements and in parallel consider demographic variations (e.g. sex) within those ROI measurements via a classifier.


\subsection{Graph-based Imputation}
We view the $v$ regions as the nodes of a graph with $X^O$ as node features. To capture adjacency among cortical ROIs
and simplify training, we construct a sparse graph by adding an edge between two brain regions if they share a boundary on the cortical surface. This undirected graph with $v$ nodes is then encoded by an ``adjacency matrix'' $\mathbf{A} \in \mathbb{R}^{v \times v}$, where $\mathbf{A}_{ij}$ is 1 if and only if nodes $i$ and $j$ are connected. As $\mathbf{A}$ does not change across subjects, then each subject is encoded by the graph $G = <X^O, \mathbf{A}>$, whose node features are the subject-specific measurements $X^O$. 

Given a graph $G$, we aim to learn its encoding into node embeddings $h_G  \in \mathbb{R}^{v \times r}$ that is optimized for imputing missing ROI measurements and predicting the label, i.e., demographic factor sex (see Figure \ref{network}). The node embeddings are learned by a Graph Isomorphism Network (GIN) \cite{xu2018how}, which compares favorably to conventional GNNs such as GCN \cite{gcn} in capturing high-order relationships across features of neighboring ROIs \cite{xu2018how}. Each layer of a GIN learns the relationships between neighboring ROIs by first summing up the feature vectors of adjacent nodes. These new vectors are then mapped to hidden vectors via a multi-layer perceptron (MLP).
The hidden vector $h_i^k$ of a particular node $i$ at the $k$-th layer is then defined as :
\begin{equation}
    h_i^k:=\text{MLP}\left((1+\varepsilon) \cdot h_i^{k-1}+\sum_{j \in \mathcal{N}_i} h_j^{k-1}\right),
\end{equation}
where $\mathcal{N}_i$ denotes nodes adjacent to node $i$ (according to $\mathbf{A}$) and the weight $\varepsilon$ of a node compared to its neighbors is learned. 

The node embeddings of the last layer $h_G:=\{h_i\}_{i\in v}$ are then fed into a graph decoder, which again is a GIN. The decoder is trained to reconstruct the missing measurements $X^M$ using $h_G$ obtained from ``shared measurements'' $X^O$ by deriving the mapping function $f(\cdot) $ so that the predicted value $\widehat{X}^{M} := f(h_{G})$ minimizes the loss function 
\begin{equation}
\mathcal{L}_{imp} := \left\|X^M-f(h_G)\right\|^{2}, 
\end{equation}
where $\| \cdot \|$ is the Euclidean distance. 

\subsection{Demographic Aware Graph-based Imputation}
As mentioned, we implement a classifier in parallel to the graph decoder (Figure \ref{network}). Given the subject-specific node embedding $h_G$ and label $y_G$ (e.g., female or male), this classifier aims to learn a function $g(\cdot)$ that maps the node embeddings of $G$ to label $y_G$, i.e., $\widehat{y}_G:=g(h_G)$. As shown in Figure \ref{network}, our model first applies a global mean pooling operation to $h_G$ in order to extract the graph embedding required for the MLP to perform the classification \cite{grattarola2022understanding}.  The loss function optimized by the classifier is then
\begin{equation}
\mathcal{L}_{cls} := y_G \log \left(g(h_G)\right)  + (1-y_G) \log \left(1-g(h_G) \right).
\end{equation}
To minimize this loss, the node embeddings $h_G$ are optimized with respect to representing demographic differences. 

Explicitly accounting for demographic differences then improves the accuracy of the imputation task as the demographic factors (i.e., sex) estimated by the classifier provide additional information further constraining the search space. Thus, the overall loss function minimized by DAGI combines imputation and classification loss, i.e.,  
\begin{equation}
    \mathcal{L}_{total} := \mathcal{L}_{imp} + \mathcal{L}_{cls}.
\end{equation}

\subsection{Implementation}
We implement the model in PyTorch using the Adam optimizer with a learning rate of 0.01. The batch size is set to 32 and the number of epochs is 300.  The dimension of node embedding $r$ is 32. Our graph encoder is composed of two GIN layers, each containing an MLP with two fully-connected layers. Our graph decoder contains one GIN layer with four fully-connected layers. Following each GIN layer, we apply ReLU functions and batch normalization to enhance stability. Codes will be available at \hyperref[https://github.com/Wangyixinxin/DAGI]{https://github.com/Wangyixinxin/DAGI}

%
%
\section{Experimental Results}
In this section, we evaluate DAGI on the NCANDA and ABCD data sets (described in Section \ref{sec3.1}). On NCANDA (Section \ref{sec3.2}), we determine the accuracy of the imputed measurements by our and other approaches by comparing them with real measurements via 5-fold cross-validation. We highlight the crucial role of explicitly accounting for the relationship between ROIs and the demographic factor sex in the imputation process by visualizing the learned embeddings and examining the discrepancy in the imputed measurements across the sexes. In an out-of-sample test on ABCD (Section \ref{sec3.3}), the curvature scores are not provided so we infer the accuracy from a classifier identifying sex just based on ABCD measurements, by including also our imputed ones, and by adding those imputed by alternative approaches that combine NCANDA and ABCD dataset in the training process. 

\begin{table}[!t]
\centering
\caption{Imputation accuracy based on 5-fold cross-validation on NCANDA. GI refers to the implementation of DAGI without the classifier. The best results are shown in \textbf{bold}. Compared to DAGI, all error scores are significantly higher ($p \leq 0.05$ based on two-sided paired t-test) with the exception of the MSE and MAE associated with the mean curvature scores produced by GIN.}
\resizebox{\textwidth}{!}{
\begin{tabular}{llcccccc}
\toprule[1.3pt]
& & \multicolumn{3}{c}{Mean Curvature } & \multicolumn{3}{c}{Gaussian Curvature} \\ \cmidrule(lr){3-5} \cmidrule(lr){6-8}
& & MSE & ~MAE & MRE & MSE & ~MAE & MRE \\ 
& & (e$^{-3}$) & (e$^{-2}$) & & (e$^{-4}$)& (e$^{-2}$) & \\\hline
\multicolumn{2}{l}{Linear Regression} \cite{montgomery2021introduction} \\
~~~~& Direct & 9.40 & 2.95 & 40.36 & 3.15 & 1.63 & 15.68 \\
& ROI-based & 8.52 & 2.12 & 31.77 & 2.24 & 1.12 & ~9.58 \\ \hline
\multicolumn{2}{l}{Multi-layer Perceptron} \cite{sklearn_api} & 8.89 & 2.56 & 35.65 & 2.99 & 1.58 & 12.90 \\ \hline
\multicolumn{2}{l}{GI} \\
& GCN \cite{gcn} & 9.80 & 3.01 & 45.29 & 3.05 & 1.60 & 14.51 \\
& GIN \cite{xu2018how} & 7.87 & 1.99 & 28.65 & 1.88 & 1.05 & ~7.22 \\
\multicolumn{2}{l}{DAGI (Proposed)} & \textbf{7.71} & \textbf{1.92} & \textbf{26.77} & \textbf{1.19} & \textbf{0.81} & \textbf{~5.41} \\
\bottomrule[1.3pt]
\end{tabular}}
\label{imputation_results}
\end{table}
\subsection{Dataset}
\label{sec3.1}
We utilize two publicly available datasets to evaluate our proposed model. The first data set (Release: NCANDA$\_$PUBLIC$\_$BASE$\_$STRUCTURAL$\_$V01 \cite{pfefferbaum2016adolescent}) consists of baseline Freesurfer measurements of all 540 participants (270 females and 270 males) of NCANDA \cite{ncanda} that are between the ages 12-18 years and report no-to-low alcohol drinking in the past year. The Freesurfer score for each of the 34 bilateral cortical regions defined according to the Desikan-Killiany Atlas \cite{desikan2006automated} consists of 5 regional measurements: average thickness, surface area, gray matter volume, mean curvature, and Gaussian curvature. The second public data release is the Data Release 4.0 of ABCD dataset \cite{casey2018adolescent}, from which we use data from all 3760 adolescents (1682 females and 2078 males) collected between ages 12 to 13.8 years for our analysis. In addition to the average thickness, surface area and gray matter volume, ABCD released the ``sulcal depth'' but does not contain the two curvature scores released by NCANDA. Imputing those curvature scores from the three shared ones is the goal here. 



\subsection{Experiments on NCANDA}
\label{sec3.2}
\subsubsection{Quantitative Comparison:}
In NCANDA, we measure the accuracy of our imputed measurements by performing 5-fold cross-validation and then record for each measurement type the average Mean Squared Error (MSE) and Mean Absolute Error (MAE) across all subjects. Based on MAE, we also compute the Mean Relative Error (MRE) to have an error score that is indifferent to the scale of the inferred measurements. To put those accuracy scores into context, we repeat the 5-fold cross-validation for other approaches. Specifically, we impute the measurements via an MLP \cite{sklearn_api} and a linear regression model \cite{montgomery2021introduction} (a.k.a., direct linear regression). As not all measurements across ROIs necessarily have a direct relationship with one another, the ``ROI-based Linear Regression'' separately fits a linear model to each ROI so that it imputes missing measurements as the linear combinations of observed measures within each individual region. We investigate our modeling choices by imputing scores without the classifier (referring to as Graph Imputation, or GI) and by replacing the GIN with a GCN \cite{gcn}. We apply two-sided paired t-tests between the error scores recorded for the proposed DAGI and each alternative approach and label p-values $\leq$ 0.05 as being significantly different. 

According to Table \ref{imputation_results}, the two approaches oblivious to ROIs, i.e., linear regression and MLP, received relatively high error scores indicating the importance of accounting for ROI-specific characteristics in the imputation process. This observation is further supported as their error scores are significantly higher (p$<$0.0017 across all scores) than those of the ROI-based linear regression. Significantly lower MRE scores than the ROI-based linear regression are recorded for GIN (p$<$0.0001), which supports our choice for encoding adjacency between ROIs in a graph structure. This encoding of the graph structure is significantly more accurate (p$<$0.0085 across all scores) than the alternative based on the GCN model. The MRE is further significantly reduced (p$<$0.0001) by guiding the training of the imputation model using the sex classifier, i.e., produced by DAGI. In summary, DAGI reported the lowest error scores across all metrics, which supports our modeling choices.   

\noindent\textbf{The Importance of the Classifier for Imputation:}
\begin{figure}[!t]
\includegraphics[width=\textwidth]{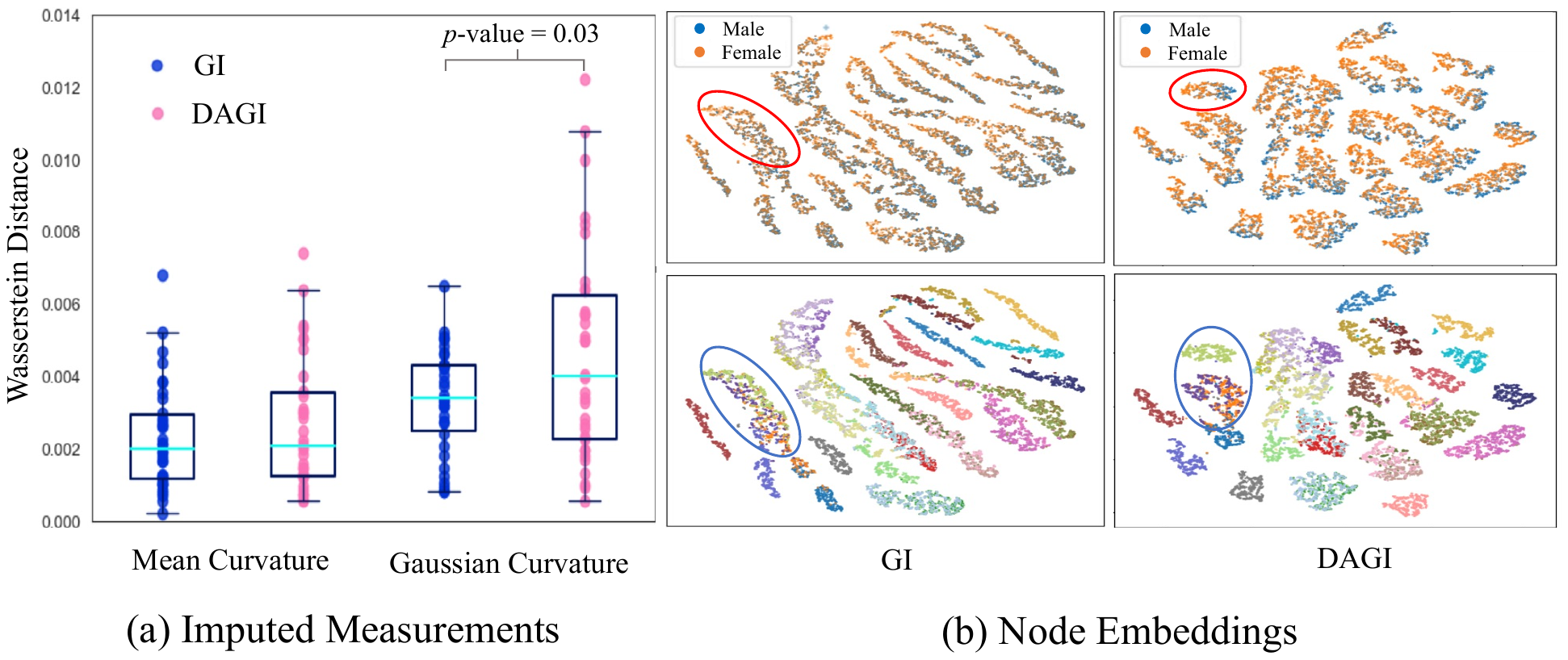}
\centering
\caption{The importance of the classifier for imputation. (a) Wasserstein distance between sexes with respect to imputed ROI curvature scores. The distances are higher for DAGI (vs. GI) and that difference is significant with respect to the Gaussian Curvature according to the two-sided paired t-test; (b) t-SNE visualization of node embeddings color-coded by sex (first row) and by ROIs (second row). Embeddings of DAGI (right column) have clearer sex differences (e.g., highlighted by red circles) and larger separation between ROIs (e.g., blue circles) compared to embeddings of GI.}
\label{fig:wd}
\end{figure}
To gain a deeper understanding of the importance of modeling demographic factors (i.e., sex) for imputing each curvature score, Figure \ref{fig:wd} (a) plots the Wasserstein distance \cite{villani2009optimal} between the sex-specific distributions of the imputed measurements for DAGI  and GI (i.e., DAGI with GIN and without classifier). We choose the Wasserstein distance as it is a fairly robust metric that ignores outliers by comparing the overall shape of distributions. While for both curvature scores the distance for DAGI is higher for the majority of ROIs (20 out of 34 ROIs for ``mean curvature'' and 19 ROIs for ``Gaussian curvature''), the difference compared to GIN across all regions is significant (p = 0.03, two-sided paired t-test) only with respect to the ``Gaussian curvature''. This finding supports that sex is important for imputations for both curvature scores but more so for the  ``Gaussian curvature'', which would also explain why in Table 1 all error scores of DAGI are significantly lower for this curvature score (than GI) but for the mean curvature it is only the MRE that is significantly lower.

\noindent\textbf{Visualizing the Node Embeddings: } Next we investigate the importance of modeling sex and ROI adjacency for the imputation task by visualizing the node embeddings of the two implementations. Shown in Figure \ref{fig:wd} (b) are the t-SNE plots \cite{van2008visualizing} of those embeddings, where each dot represents an imputed ROI measurement of an NCANDA subject and in the top row the color refers to a specific ROI. While the embeddings by GI are clearly separated by region (Figure \ref{fig:wd} (b) left, first row), they fail to distinguish measurements by sex, i.e., blue and orange dots overlap with each other (Figure \ref{fig:wd} (b) left, second row). Our approach, (Figure \ref{fig:wd} (b) right), effectively distinguishes the sexes in the latent space (first row) while also keeping separate clusters for the ROIs (second row) as highlighted by the red circles. This separation is important for imputing the ROI measurements according the to error scores reported in Table 1.

\noindent\textbf{Visualizing the Brain Graph:} 
\begin{figure}[!t] 
\centering
\includegraphics[width=1\textwidth]{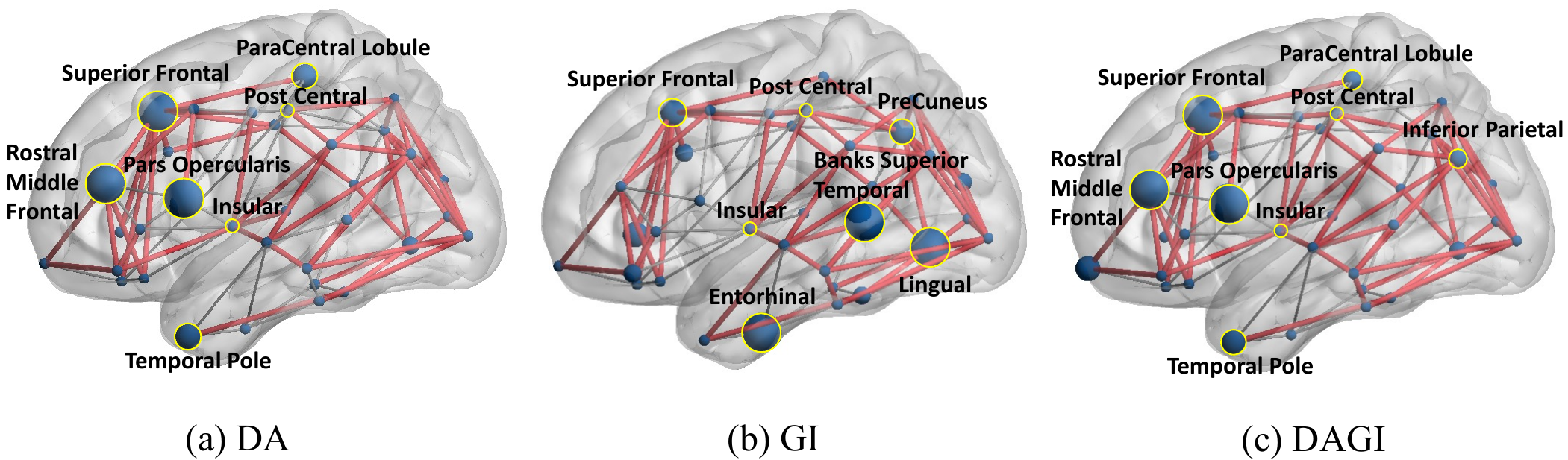}
\caption{Graph node and edge importance according to GNNExplainer \cite{ying2019gnnexplainer}. Each node corresponds to an ROI. Larger nodes represent higher contributions with the most influential ones highlighted by a yellow circle.  Red edges are those of the subgraph deemed most important for the task at hand. According to the figure, individual ROIs are more important for sex classification ((a) and (c)), while the relationship between ROIs is more important for imputation ((b) and (c)).}\label{fig2}
\end{figure}
We investigate the importance of specific ROIs in imputing the measurements by visualizing the graph structure via the GNNExplainer \cite{ying2019gnnexplainer}. GNNExplainer defines the subgraph most important for the task at hand as the one whose predicted distribution maximizes the mutual information with the one derived from the original graph. Figure \ref{fig2} visualizes this subgraph with red edges (i.e., the connection between ROIs). The importance of individual nodes (i.e., ROI) is encoded by their radius. It is striking that the subgraph of DAGI (Figure \ref{fig2} (c)) is a combination of the graphs of the other two models, i.e., the importance of nodes is similar to those of the approach with solely Demographic Aware module, referring to as DA (Figure \ref{fig2} (a)) while the importance of edges agrees with the model that only relies on the imputation model, i.e., GI in Figure \ref{fig2} (b). This suggests that individual ROIs are more important for classification while the interaction between ROIs is more important for imputation. Based on those plots, we conclude that identifying sex is mostly driven by pars opercularis, rostral middle frontal, and superior frontal regions, which is in line with the literature \cite{lv2010gender,luders2006gender}. However, imputation heavily relies on the interaction between neighboring regions (such as between post central and insula regions).


\subsection{Out-of-sample Test on ABCD.}
\label{sec3.3}
Using DAGI trained on NCANDA (i.e., the most accurate model according to Table 1), we now impute the missing curvature scores on ABCD. Given the lack of ``ground truth'' with respect to the missing ABCD measurements, we indirectly evaluate the quality of the imputed values by comparing the accuracy of a classifier identifying sex on the 3760 ABCD participants with and without utilizing the imputed measurements. This experimental setup is based on the observation that if the imputed measurements are accurate then they should hold pertinent and discriminatory details that could be utilized for downstream tasks, such as sex classification. 

The sex classifier is a three-layer MLP model, whose balanced test accuracy is measured via 5-fold cross-validation. In order to remove the confounding effect of brain size on sex classification, we normalize the  ``average thickness'', ``surface area'' and  ``gray matter volume'' measurements by the supratentorial volume \cite{o2009encyclopedia,pfefferbaum2016adolescent}. Note, the imputed curvature scores are left unchanged since they are not confounded by brain size as their Pearson correlation \cite{cohen2009pearson} with the supratentorial volume is insignificant for all regions (maximum correlation is 0.053, p$<$0.01). 

\begin{table}[!t]
    \centering
    \caption{Balanced accuracy of an MLP classifying sex based on ABCD with and without imputed brain measurements. The best results are shown in \textbf{bold}.  All accuracies are significantly lower than DAGI (p-value $\leq 0.01$ according to McNemar's test). }
    \resizebox{1\textwidth}{!}{
    \begin{tabular}{lll}
    \toprule[1.3pt]
      \multicolumn{2}{l}{Measurements Used by Classifier} &  Accuracy \\ \toprule[0.8pt]
      \multicolumn{2}{l}{~~Only ABCD scores} &  ~~0.838 \\ \cline{1-3}
      \multicolumn{1}{l}{\multirow{3}[1]{*}{~~Including imputed curvature scores~}}  & MICE \cite{van2011mice} (trained on NCANDA $\&$ ABCD)~~~ &  ~~0.811 \\ \cline{2-3}
       & GINN \cite{SPINELLI2020249} (trained on NCANDA $\&$ ABCD) &   ~~0.832 \\ \cline{2-3}
       & DAGI (trained on NCANDA only) &   ~~\textbf{0.845} \\ 
        \bottomrule[1.3pt]
    \end{tabular}}
    \label{abcd_verification}
\end{table}

According to Table \ref{abcd_verification}, the balanced accuracy of the classifier just on the ABCD measurements is 83.8 $\%$, which then significantly improves (p=0.008, McNemar's test \cite{mcnemar1947note}) to 84.5 $\%$ once the imputed scores are added. To put the improvement into context, we also record the classification accuracy with respect to curvature scores generated by the traditional imputation methods MICE \cite{van2011mice} and the deep learning-based GINN \cite{SPINELLI2020249}. Since these methods are originally designed for randomly missing values (Figure \ref{bg} (a)) and thus cannot work on the ABCD dataset alone, we train them to impute missing values on matrices containing both the NCANDA and ABCD measurements. 
Surprisingly, the inclusion of the curvature measurements imputed by MICE and GINN results in significantly lower classification accuracy than DAGI (p$<$0.01, McNemar's test). The accuracy is even worse than the classifier solely based on ABCD scores. This suggests that they fail to accurately impute the curvature scores and instead mislead the classifier by making the data more noisy. This might be attributed to the fact that these methods are typically designed for randomly distributed missing values, and thus may not be suitable for our specific scenario where specific measurements are entirely missing in a dataset (Figure \ref{bg} (b)). For this scenario, the significant improvement achieved via the curvature scores predicted by DAGI demonstrates the utility of imputing brain measurements for enhancing downstream tasks. 


\section{Conclusion}
The accuracy of classifiers (e.g. identifying sex from brain ROI measurements) applied to publicly available data can be negatively impacted by the absence of entire measurements from that data set. Instead of imputing the scores by merging the data set with ones that contain the measurements, we propose to rephrase the problem as a prediction task in which we learn to predict missing measurements from those that are shared across data sets.  We do so by coupling a graph neural network capturing the relationship between brain regions and a classifier to model demographic differences in ROI brain measurements. Compared to existing technology, our proposed method is significantly more accurate in imputing curvature scores on NCANDA. Imputing the measurements on ABCD and then feeding them into a classifier also result in more accurate sex identification than solely relying on the ROI measurements provided by ABCD. Overall, our framework provides a novel and effective approach for imputing missing measurements across data sets as it is only trained once on the data set that contains the values. This might also have important implications for generalizing neuroscientific findings of deep learning approach across data sets as they could now rely on the same set of measurements. 

\subsubsection{Acknowledgments} This work was partly supported by funding from the National Institute of Health (DA057567,  AA021697, AA017347, AA010723, AA005965, and AA028840), the DGIST R\&D program of the Ministry of Science and ICT of KOREA (22-KUJoint-02), Stanford School of Medicine Department of Psychiatry and Behavioral Sciences Faculty Development and Leadership Award, and by the Stanford HAI Google Cloud Credit.

%
%
%
\bibliographystyle{splncs04}
\bibliography{ref}

\end{document}